\documentclass[runningheads]{llncs}
\usepackage{cite}
\usepackage{hyperref}
\usepackage[table,xcdraw]{xcolor}
\usepackage{multirow}
\usepackage{tabularx}
\usepackage{float}
\usepackage{amsmath}    
\usepackage{amssymb}    

\usepackage{graphicx}
\begin{document}
\title{Identifying Gender Stereotypes and Biases in Automated Translation from English to Italian using Similarity Networks}
%
%
\author{Fatemeh Mohammadi\inst{1}\orcidID{0000-0002-3100-7569} \and
Marta Annamaria Tamborini\inst{2}\orcidID{0009-0005-6904-3708} \and
Paolo Ceravolo\inst{1}\orcidID{0000-0002-4519-0173}\and
Costanza Nardocci\inst{2}\orcidID{0000-0002-0138-0581}\and
Samira Maghool\inst{1}\orcidID{0000-0001-8310-2050}
}
\authorrunning{F. Mohammadi \and et al.}
%
\institute{
  Department of Computer Science
  \and 
  Department of Italian and Supranational Public Law \\Università degli Studi di Milano, Milan, Italy
}

\maketitle              
\begin{abstract}
This paper is a collaborative effort between Linguistics, Law, and Computer Science to evaluate stereotypes and biases in automated translation systems. We advocate gender-neutral translation as a means to promote gender inclusion and improve the objectivity of machine translation. Our approach focuses on identifying gender bias in English-to-Italian translations. 
First, we define gender bias following human rights law and linguistics literature. Then we proceed by identifying gender-specific terms such as \textit{she/lei} and \textit{he/lui} as key elements. We then evaluate the cosine similarity between these target terms and others in the dataset to reveal the model's perception of semantic relations. Using numerical features, we effectively evaluate the intensity and direction of the bias. Our findings provide tangible insights for developing and training gender-neutral translation algorithms. 
 
\end{abstract}
\section{Introduction}
Recent advances in Natural Language Processing (NLP) have marked significant milestones, notably with the emergence of Large Language Models (LLMs), which have led to undeniable performance improvements in various NLP tasks. Among these tasks, Machine Translation (MT) is one of the most widely used yet critical applications ~\cite{Pikuliak2023}. The integration of Artificial intelligence (AI) into translation services has undoubtedly facilitated cross-cultural communication and nowadays automated translation providers are widely used in every industry. For example, today Google Translate, the multilingual neural machine translation service developed by Google, has over 610 million users per day. Despite the undeniable usefulness of such technologies, the existence of gender biases and stereotypes in MT is a fact~\cite{stanczak2021survey}. 

The concept of gender bias takes on a multifaceted nature depending on the field of study. Within legal and human rights frameworks, it signifies the unfair treatment of individuals based on their gender. This can encompass discriminatory laws, biased legal processes, and the resulting human rights violations~\cite{feng2019gender}, ~\cite{undp2019genderequality}. Conversely, linguistics examines how language itself perpetuates societal inequalities through gender bias. This manifests in grammatical structures that favor one gender, vocabulary lacking female equivalents for certain roles, and the way words carry specific connotations about masculinity and femininity ~\cite{lakoff1975language}; ~\cite{mullenbrock2011gender}; ~\cite{cameron1998presumptive}. Recognizing these distinct yet interrelated aspects of gender bias is essential to tackling its presence in both legal systems and our daily conversations.

This paper focuses mainly on gender bias defined as the preference or prejudice for one gender over the other~\cite{stanczak2021survey}. The influence of gender stereotypes and prejudices in language is well-documented in Italian and English linguistic studies. Research lines in both law and linguistics have shown ~\cite{beeghly2021stereotyping, DAmico2023, morris1989stereotypic, alexander1992what} that there is a strong link between the spilling of stereotypes in language and direct or indirect discrimination~\cite{coe2018handbook, doughman-etal-2021-gender}. It is thus crucial to properly identify and correct stereotypes and biases in MT translations. Language is both a reflection and a shaper of societal norms~\cite{doughman-etal-2021-gender} and as such, automated translation systems that rely on large corpora of text inherit and amplify these biases~\cite{DAmico2023}. It is therefore important to quantify and understand language biases, as such biases can reinforce the psychological status of different groups. 

The multidisciplinary approach we develop involves the definition of the legal categories behind gender stereotypes, the linguistic factors that contribute to such biases, and the technical aspects within computer science that can be used to quantify these biases. Since translations are not only the transposition of words but also the transmission of cultural connotations and social norms, our goal is to study the manifestation of gender bias in Italian translations of English texts. Our methodology builds on existing approaches \cite{bolukbasi2016man, bolukbasi2016quantifying} but enhances the ability to detect the intensity and direction of bias through specially designed features.

Based on the above goal, the research question underlying this contribution has been to study how English to Italian gender language is addressed by MT by attempting to answer the research questions.
(i) \textit{How does MT from English to Italian affect the gender of language?}
(ii) \textit{Is the effect of MT related to bias in the algorithms or in the source data used to train MT?}
(iii) \textit{Can we precisely identify the intensity and direction of the MT bias for each term?}

More in general, this paper is positioned in the ``notional vs. grammatical'' gender languages debate and this analysis could be repeated for multiple notional/grammatical-language pairs.
Our research began by defining gender bias categories and definitions and started with the collection of word corpora containing both gender-specific and gender-neutral terms. These words were represented in a feature space and organized into a similarity network to compute their similarity. Through this process, we were able to identify similarities with gender-specific terms such as \textit{she/lei} and \textit{he/lui}. We used these similarities to derive features that capture the intensity and direction of the bias. Furthermore, we conducted a comparative analysis between the bias produced by FastText as a word embedding method and Google Translate as an MT model when translating text from English to Italian. Our results showed a significant shift in the direction of bias after translation. Furthermore, our methodology effectively identified terms susceptible to gender bias, providing valuable insights for the development of gender-neutral MT algorithms.

By advocating for more inclusive and unbiased translations, we aim to contribute to a fairer representation of gender in automated translation systems. More specifically, the paper is organized as follows: Section 2 provides background and related work, Section 3 details our methodology, Section 4 presents our results, and Section 5 ends the paper with concluding remarks.

\section{Background and Related Work}\label{s:rw}
\subsection{Gender Bias and Discrimination in Law and Linguistics}
This work contributes to the broader discourse on AI and discrimination which is of the utmost relevance since Europe is currently setting its legal framework on AI. Specifically, on 13 March 2024, the European Union enacted its first comprehensive regulation to address these concerns~\cite{EU2024} and on 17 May 2024 the Council of Europe adopted the first-ever international legally binding treaty aimed at ensuring the respect of human rights, the rule of law and democracy legal standards in the use of artificial intelligence (AI) systems ~\cite{CouncilofEurope2024}. Concerning potential unequal treatment leading to discrimination, the inherent risk of AI models lies in their potential to perpetuate discrimination against minority groups due to biases in the data sets or the same architecture of the system. The new European legal framework emphasizes the need for a more conscientious and responsible approach to AI design and evaluation \cite{Nardocci2021}. As a consequence, eliminating AI-induced discrimination is at the heart of current research trends \cite{maghool2023enhancing}. 

Previous works addressed this topic for other languages~\cite{bolukbasi2016man} or focused on other NLP techniques ~\cite{park2018reducing,bolukbasi2016quantifying,may2019measuring} and some tried to provide a taxonomy of gender bias in texts (in English)~\cite{doughman-etal-2021-gender}. A comprehensive review of the current legal literature on the topic \cite{Nardocci2021} outlines several stages in AI design where discrimination may manifest. 

MT based on machine learning technologies, represents an example of how AI perpetuates stereotypes typical of human language~\cite{DAmico2023} from one culture to another. Let us recall the case of Google Translate~\cite{Prates2018}, the MT system exhibited biased behavior by translating job titles from English into languages that incorporate masculine/feminine characterization. This is particularly evident when lower-income and non-leadership positions are translated with their female counterparts, while higher-income and leadership positions are associated with their male counterparts, perpetuating gender stereotypes. For example, "the nurse" would be translated with "l'infermiera" (female nurse) and "the doctor" with "il medico" (male doctor)~\cite{AdamsLoideain2019,Prates2018}.

The importance of gender in human experience is universally acknowledged, reflected in the linguistic expressions of femininity and masculinity present across languages. However, languages differ in their methods of encoding gender. English, for instance, is classified as a notional gender language, primarily conveying the gender of human referents through personal pronouns, possessive adjectives (e.g., he/him/his; she/her/hers), and gender-specific terms (e.g., man; woman). In contrast, grammatical gender languages such as Italian utilize a system of morphosyntactic agreement, where gender markers extend beyond nouns to encompass verbs, determiners, and adjectives \cite{Corbett2006}. This distinction becomes particularly significant in translation contexts\cite{garzone2020chapter}, notably when the source language lacks gender information regarding a referent and the target language operates within a grammatical gender framework and, although controversial to this day, prescribes the grammatical rule of the "inclusive masculine". This rule is embodied in the fact that if only one masculine exponent is present in a group, the plural masculine will apply. For example, in a class consisting of 10 people 9 are women but 1 is a man, it is grammatically correct for the teacher to say "Buongiorno a tutti" (-i standing for masculine plural). Furthermore, until recently, there were no feminine words to define higher professional positions. With social change and the promotion of female participation in public life, such terms have been coined (e.g. Professoressa, Dottoressa, Avvocata, Ingegnera, etc.) but still part of public opinion is skeptical to use such terms~\cite{Robustelli2022} to the point that in 2023 Italy elected its first female prime minister who decided to be addressed by "Signor Presidente" (Mr. President) as a political statement~\cite{DAmico2023}. Italian linguistics studies nowadays converge on the assumption that Italian is a sexist language. The essay "Sexism in the Italian Language" \cite{Sabatini1987}  has been the first seed of a debate that bloomed much later. Currently, the Italian language is changing because much has changed in the role of Italian women within society. However, this change is not organic, structured, or systematic as it should be in a society that proactively strives for new relationships between women and men~\cite{SulisGheno2022}. 

Recent research in computation and language~\cite{Piergentili2023} advocated for gender-neutral translation (GNT) as both a manifestation of gender inclusion and an objective for MT models. Our research aligns with and contributes to this perspective by proposing a methodology for identifying terms that exhibit gender bias in MT. Similar studies have been applied to English~\cite{bolukbasi2016man}, to Sentiment Analysis~\cite{park2018reducing}, to word embedding methods~\cite{bolukbasi2016quantifying,may2019measuring} but never to the English to Italian translation. Through our findings, we provide tangible insights to inform the development and training of GNT algorithms, thus promoting more inclusive and unbiased translations.

\subsection{Gender Bias and Discrimination in Computer Science}

We were inspired by the research of \cite{bolukbasi2016man}. They created some analogies like King: Queen using Google Word2Vec and then asked human annotators to rate them as biased/appropriate. To detect bias, they used cosine similarity to measure the similarity of analogies with she-he. They showed that word embeddings contain biases in their geometry that reflect gender stereotypes in the wider society. 

In another paper by \cite{Stanovsky2019-hs}, they composed a challenge set for gender bias in MT called WinoMT, which contains 3,888 instances and is balanced between male and female, and between stereotypical and non-stereotypical gender roles (e.g. a female doctor versus a female nurse). They then translated them into four different language categories (Romance, Slavic, Semitic, and Germanic) using six widely used MT models representing the state of the art in commercial and academic research, such as Google Translate. They showed that MT models significantly tend to translate based on gender stereotypes rather than more meaningful contexts.

Another related study is the one by \cite{Biasion2020-yb}. They looked at gender bias in Italian word embeddings. They made a list of gender definition pairs: [lui (he), lei (she)] and then calculated the vector difference like \(\mathbf{lui}\)-\(\mathbf{lei}\) to get the direction of the bias. They also used cosine similarity to measure the differential association between target and attribute word sets. They used FastText as the word embedding method and the target set of their work consisted only of occupations in Italian. 

In another paper by \cite{Prates2020-dc}, they take an extensive list of job titles and construct some sentences like "He/She is an Engineer" (where "Engineer" is replaced by the job title of interest) in 12 different gender-neutral languages such as Hungarian, Chinese, Yoruba, and several others. They then translate these sentences into English using the Google Translate API and collect statistics on the frequency of female, male, and gender-neutral pronouns in the translated output. We then show that Google Translate has a strong bias towards male pronouns, especially in fields typically associated with gender imbalance or stereotypes, such as STEM (Science, Technology, Engineering, and Mathematics) jobs.

Finally, the literature suggests that most word embedding models, such as Word2Vec and FastText, have a gender bias. This bias can influence how models learn patterns, potentially reinforcing societal biases and stereotypes. Consequently, AI and ML models, including machine translation (MT) models, are also susceptible to such biases in word embedding. In this research, we aim to investigate the extent to which this bias is caused by word embedding or translation, and whether translation affects the intensity and nature of this gender bias.

\section{Material and Methods}
Our approach is designed to uncover gender bias in translations from English to Italian. We do this by computing similarity scores between certain gender-specific target words, such as \textit{she/he}, and the other words in a word list that combine gender-neutral and gender-specific words. In the following sections, we will explain our procedure by breaking it down into its main stages.

\subsection{Data collection}
As part of our effort to highlight gender bias in automated translation, we seek a collection of words that include various gender-specific and gender-neutral terms. For this purpose, we used the analogies generated by word embedding provided by~\cite{bolukbasi2016man} in the Appendix section of the paper. The authors generated 236 (word pairs) analogies by an analogy generator which gets a seed pair of words (a,b) determining a seed direction (\(\mathbf{a}\)-  \(\mathbf{b}\)) corresponding to the normalized difference between the two seed words. An example of analogy in this paper is she:sewing:: he: carpentry. It represents a relationship between words based on their contextual associations. It means that there is a strong association of "she" with "sewing" and "he" with "carpentry." This analogy reflects gender stereotypes present in the training data. Because Word embedding captures patterns and associations in the text they are trained on, including biases. 

\subsection{MT reference}
Given the widespread use of \textit{Google Translate}, we included its API in our study. Bias in such a widely used tool raises serious concerns and underscores the importance of addressing it.

\subsection{Pre-processing and Word Embedding}

To measure similarity, we first took the list of analogies and broke it down into single words (472 words in total). Our approach in using single words (detached from the grammatical or contextual environment) is similar to the WEAT or Verb Extraction approaches \cite{jentzsch2019semantics}. First, we removed duplicate words (27 removed). Also, some words had plural and singular forms. Since we use similarity scores and there is no significant difference between the plural and singular forms of a word, we decided to keep only the singular form of words, resulting in a list of 333 words. This ensures that the text data is standardized and ready for the similarity measurement algorithms. 

After preparing the list of words, we need to organize words in a vector space to compute similarity scores. Given the requirement to analyze both English and Italian texts and to perform a comparative assessment, we chose a multilingual approach to word embedding so we selected FastText. \textit{FastText} \cite{Kuyumcu2019} is a word embedding method that uses a vectorization process by considering subwords (N-grams) as the smallest unit instead of single words. This approach makes it independent of the distribution of words in a vocabulary and allows generalizing across languages, facilitating the transfer of knowledge learned in one language to another. In contrast, other options, such as \textit{Word2Vec}, are inherently language-dependent and thus unsuitable for our research design.

\subsection{Similarity measurements}

We utilize a graph-based approach to analyze the relationships between data points. This approach enables us to quantify the likeness between data points, facilitating the construction of a weighted network denoted as $N = (V, E)$. Here, $V$ signifies the nodes (vertices) within the network, while $E$ denotes the links (edges) connecting them \cite{maghool2023enhancing}. In this application, each node corresponds to a word, and the weights of the edges reflect the degree of similarity between these words. 

To calculate the similarity (proximity) between two nodes, we use the cosine similarity function applied to the embedded vectors of the given nodes \(i\) and \(j\) in the network. This similarity is computed using the following formula~\cite{Bellandi2020-uy}:  

\[
s(i, j) = \frac{\mathbf{x}_i \cdot \mathbf{x}_j}{\lVert \mathbf{x}_i \rVert \lVert \mathbf{x}_j \rVert}
\]

We have generated a robust similarity network by employing cosine similarity measurements, particularly with gender-specific terms such as she/he and lei/Lui. Before choosing \textit{he/lui} and \textit{she/lei} as target words we measured the internal similarity between them. i.e. we measured the similarity between she and he (0.61) and the similarity between lui and lei (0.85). so these target words are quite different so we can rely on them as a good differentiated point for identifying bias. Also note that we could have used other words, e.g. woman and man, as the
gender-pair in the task. We chose she and he because they are frequent and do not have fewer alternative word senses (e.g., man can also refer to mankind) \cite{bolukbasi2016man}.

Because of its efficiency, cosine similarity is chosen for comparing vectors such as word embeddings. Unlike measures based solely on magnitude, cosine similarity evaluates the angle between vectors, emphasizing their directional alignment. This property is advantageous when comparing word vectors, as it prioritizes vector direction over absolute values and provides more meaningful similarity judgments.

\subsection{Detecting gender bias}
To study the manifestation of gender stereotypes or biases, one approach is to quantify the proximity of words to gender-specific terms such as \textit{he/lui} vs. \textit{she/lei}~\cite{bolukbasi2016man}. To enhance the accuracy of bias identification, we introduce a numerical feature that measures the absolute difference between the similarity to \textit{she} and the similarity to \textit{he}. This feature, called \textit{GenderBiasIntensity}, quantifies the intensity of bias without considering its direction at this stage. If there are two target words with different genders, denoted as \(\mathbf{X}\) and \(\mathbf{Y}\), the gender bias intensity of \(\mathbf{w}\) can be expressed as follows~\cite{jentzsch2019semantics}:  

\[
\text{GenderBiasIntensity} (\mathbf{w}) = \left| \cos(\mathbf{w}, \mathbf{X}) - \cos(\mathbf{w}, \mathbf{Y}) \right|
\]

So for determining the gender bias of a specific word, after tokenization and word embedding, we compute the cosine similarity of that word with these target words. For example, if the similarity score between \textit{she} and \textit{doctor} is lower than the similarity between \textit{he} and \textit{doctor}, this suggests a gender bias against women since the model assumes that a \textit{doctor} is predominantly male.

To determine the direction of bias, we introduce another feature called \textit{GenderBiasDirection}, which represents the difference between similarity to \textit{she} and similarity to \textit{he} (without taking the absolute value). A positive value for this feature suggests a bias towards women, whereas a negative value indicates a bias towards men. Based on this definition, the gender bias direction of \(\mathbf{w}\) can be expressed as follows:

\[
\text{GenderBiasDirection} (\mathbf{w}) = \cos(\mathbf{w}, \mathbf{X}) - \cos(\mathbf{w}, \mathbf{Y})
\]

To examine the extent of post-translation similarity changes, we introduce another numerical feature that adopts a comparative approach. This feature is calculated by subtracting the similarity to \textit{she} (the target word in the source language, i.e., English) from the similarity to \textit{lei} (the target word in the destination language, i.e., Italian), represented as \(\mathbf{lei} - \mathbf{she}\). This indicates whether similarity scores increase or decrease after translation. Similarly, we can perform the same operation for \textit{lui} and \textit{he}, expressed as \(\mathbf{lui} - \mathbf{he}\). Based on this definition, this feature can be calculated using the following formula:

\[
\text{PostTranslationSimilarityChanges} (\mathbf{w}) = \cos(\mathbf{w}, \mathbf{X}_{\text{dest}}) - \cos(\mathbf{w}, \mathbf{X}_{\text{src}})
\]

\section{Discussion}
In this section, we briefly describe the similarity measurement and the gender bias analysis we did, our main results, and the final discussion and comparison of our work with related ones.

\subsection{Similarity Scores Analysis}
First of all, we need to calculate the similarity of English and Italian-translated words with our target terms. Here the results of these measurements were presented.

\subsubsection{Similarity Scores in English Words}
As mentioned in the methodology section, we compute the cosine similarity between some selected English words and two target terms (\textit{he/she}). Table 1 shows some of these calculated similarities. For some certain words such as \textit{nurse} which are highlighted in red, there is a significant difference in the similarity scores (i.e the difference is more than 0.1) with \textit{he} and \textit{she}, in comparison with other words in the table, which indicates a gender bias in the word embedding method, i.e. FastText. This observation, which suggests bias in word embeddings, has also been confirmed by other studies, such as~\cite{bolukbasi2016quantifying,may2019measuring}.

\begin{table}[ht]
\centering
\caption{Cosine similarity with He and She for some English words.}
\label{tab:1}
\begin{tabular}{|c|c|c|}
\hline
\textbf{English Words} & \textbf{Similarity with She} & \textbf{Similarity with He} \\ \hline
architect              & \textbf{0,211}               & \textbf{0,065}              \\ \hline
doctor                 & 0,119                        & 0,137                       \\ \hline
liberal                & 0,193                        & 0,209                       \\ \hline
nurse                  & \textbf{0,556}               & \textbf{0,241}              \\ \hline
politician             & \textbf{0,301}               & \textbf{0,413}              \\ \hline
salespeople            & 0,130                        & 0,178                       \\ \hline
socialists             & \textbf{0,092}               & \textbf{0,249}              \\ \hline
surgeon                & 0,049                        & -0,023                      \\ \hline
\end{tabular}
\end{table}

The distribution of word similarity concerning \textit{he} and \textit{she} is visualized using a scatter plot. Figure 1 shows the similarity scores projection for English words with these target terms. The x-axis represents the similarity score between words with \textit{he}, while the y-axis represents the similarity score between words with \textit{she}. Another insight from this projection onto a 2-dimensional space is the possibility of drawing a diagonal line from (0,0) to (1,1). Words that deviate more significantly from this diagonal line indicate a higher degree of bias. Dots positioned above the line indicate a bias against females, whereas dots below the line suggest a bias towards males. 
\begin{figure}[ht]
  \centering
    \includegraphics[width=0.5\textwidth]{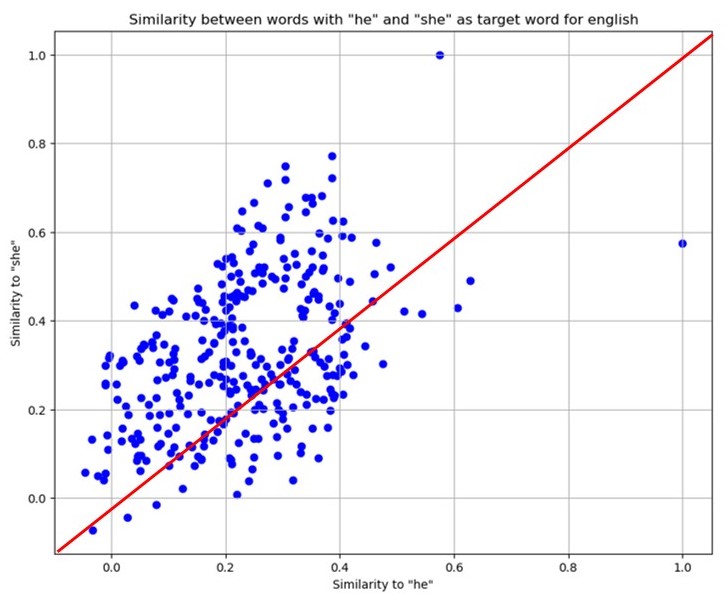}
  \caption{Scatter plot illustrating the distribution of similarity scores of English words relative to two different target words (he/she)}
\end{figure}
\subsubsection{Similarity Scores in Italian Words}
Following the methodology outlined above, we used the Google Translate model for translation to ensure the reliability of our methods and translations. We followed a similar procedure for Italian as in English to facilitate comparison between the two languages. We also use 2 gender-specific words: \textit{lei/lui}. Table 2 shows some of these calculated cosine similarities for Italian translations. As before some rows with significant differences (almost 0.1 or above) in the similarity scores with lui and lei are highlighted in red to have a comparison of similarity scores between the two languages. 

\begin{table}[ht]
\centering
\caption{Cosine similarity with \textit{Lei} and \textit{Lui} for some Italian translations.}
\label{tab:2}
\begin{tabular}{|c|c|c|}
\hline 
\textbf Italian Translation & \textbf Similarity with Lei & \textbf Similarity with Lui \\ 
\hline
architetto & \textbf{0.332} & \textbf{0.422} \\ \hline
medico     & 0.476          & 0.502          \\ \hline
liberale   & 0.341          & 0.389          \\ \hline
infermiera & \textbf{0.624} & \textbf{0.419} \\ \hline
politico   & \textbf{0.315} & \textbf{0.428} \\ \hline
venditori  & \textbf{0.127} & \textbf{0.230} \\ \hline
socialisti & 0.245          & 0.284          \\ \hline
chirurgo   & 0.468          & 0.498          \\ \hline
\end{tabular}
\end{table}

From this table, it is obvious that for most of the words, the similarity with the target word itself increases after translation. However, in some cases such as \textit{architetto, infermiera, chirurgo}, and \textit{socialisti,} the intensity of the similarity to lui/he and lei/she, decreases. We conducted an additional comparison by calculating the similarity between "architetta," the feminine version of "architetto," and our two target words. The results showed that the similarity between "architetta" and "lei" is 0.49, and between "architetta" and "lui" is 0.38. This indicates that the similarity of words to the target words is strongly influenced by how the machine translation handles the word and the gender it assigns. This preliminary conclusion will be further examined in the following section. 

Figure 2 shows the scatter plot of the cosine similarity scores calculated for Italian translations with lei and lui as target words. Comparing Figure 1 (which shows the projection of English words with she/he) with Figure 2 (which shows the projection of Italian translations with lei/lui) also confirms the previous observation of increased similarity scores after translation. For the Italian translations, most of the words fall within the range of $0.2$ to $0.6$, while for the English plot, the words are spread across the range of $0.0$ to $0.6$. But the overall distribution is almost the same for both languages. 
\begin{figure}[htbp]
  \centering
    \includegraphics[width=0.5\textwidth]{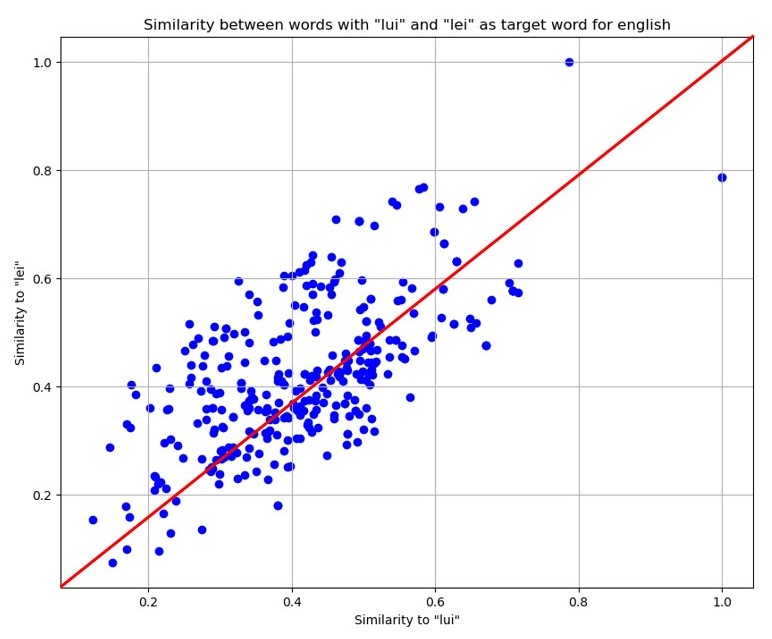}
  \caption{Scatter plot illustrating the distribution of similarity scores of Italian translations relative to two different target words (lui/lei)}
\end{figure}

We have to note that just relying on similarity scores can not lead us to the correct analysis of gender bias so in the following sections we will present some numerical features which can give us a more accurate insight about gender bias intensity and direction in English and Italian translated.

\subsection{Gender Bias Detection and Analysis}
In this section, we analyze the results obtained from calculating GenderBiasIntensity and GenderBiasDirection for both English and Italian translations. Our objective is to identify gender bias in English and examine how the translation from English to Italian can affect the intensity and direction of bias.

\subsubsection{Gender bias in English words}

After measuring the similarity scores with \textit{he} and \textit{she} we can calculate the \textit{GenderBiasIntensity} feature. This feature's range spans from about 0.0 to 0.44 for English words. Within this range for \textit{GenderBiasIntensity}, we categorize our words into five primary bins which are indicated in Table 3. Bin 1 has the lowest intensity of bias but as we advance towards bin 5, the intensity of bias increases. Table 3 displays the count of words in each bin along with representative example words for each bin. 

\begin{table}[ht]
\centering
\caption{Number of words per bin along with representative examples for comparing the intensity of bias.}
\label{tab:3}
\begin{tabular}{|c|c|c|c|}
\hline
\textbf{Bins} & \textbf{Range} & \textbf{Words Count} & \textbf{Examples}                                                                             \\ \hline
1             & 0.0 - 0.1      & 123                  & \begin{tabular}[c]{@{}c@{}}barber, baseball, brilliant,\\ briefcase, volleyball\end{tabular}  \\ \hline
2             & 0.1 - 0.2      & 95                   & \begin{tabular}[c]{@{}c@{}}architect, carpenter, charming, \\ cheerful, dictator\end{tabular} \\ \hline
3 & 0.2 - 0.3 & 78 & \begin{tabular}[c]{@{}c@{}}adorable, beautiful, cosmetics, \\ homemaker, housewife\end{tabular} \\ \hline
4             & 0.3 - 0.4      & 30                   & \begin{tabular}[c]{@{}c@{}}actress, aunt, fiance, gorgeous,\\ maternity\end{tabular}          \\ \hline
5             & 0.4 - 0.5      & 7                    & \begin{tabular}[c]{@{}c@{}}daughter, girl, he, she, \\ sister\end{tabular}                    \\ \hline
\end{tabular}
\end{table}

By a detailed examination of the words within each bin, it becomes apparent that bins 4 and 5 mainly contain gender-specific terms such as \textit{he, she, aunt, niece, sister, her, herself, }etc. Conversely, the first three bins exhibit varying degrees of gender bias, ranging from $0.0$ to $0.3$. For instance, words like \textit{housewife, handbag, homemaker, sewing, softball}, and \textit{midwife} appear in the third bin, which is recognized as female stereotypical according to a study conducted by~\cite{bolukbasi2016man}.

The range of \textit{GenderBiasDirection} spans from $-0.42$ to $0.44$ and like \textit{GenderBiasIntensity}, we can partition it into 5 bins for each gender based on the sign of the feature. Table 4 shows the number of words per bin for male and female direction.

\begin{table}[ht]
\centering
\caption{Number of words per bin for male and female direction.}
\label{tab:4}
\begin{tabular}{|cc|c|c|c|}
\hline
\multicolumn{1}{|c|}{\textbf{Bins}} & \textbf{Range} & \textbf{Female-directed} & \textbf{Male-directed} & \textbf{Total} \\ \hline
\multicolumn{1}{|c|}{1} & 0.0 - 0.1 & 67 & 56 & \textbf{123} \\ \hline
\multicolumn{1}{|c|}{2} & 0.1 - 0.2 & 57 & 38 & \textbf{95}  \\ \hline
\multicolumn{1}{|c|}{3} & 0.2 - 0.3 & 71 & 7  & \textbf{78}  \\ \hline
\multicolumn{1}{|c|}{4} & 0.3 - 0.4 & 30 & 0  & \textbf{30}  \\ \hline
\multicolumn{1}{|c|}{5} & 0.4 - 0.5 & 6  & 1  & \textbf{7}   \\ \hline
\multicolumn{2}{|c|}{\textbf{Total}}                 & \textbf{231}             & \textbf{102}           & \textbf{333}   \\ \hline
\end{tabular}
\end{table}

Bins 2 and 3 are more significant in identifying gender bias. Approximately 173 words, or 52\% of the total, show either male or female bias. This suggests a significant risk of bias in word embedding methods such as FastText, where more than half of the instances may lead to unfair representation of males and females. Another important observation is that the number of biases against females over males is greater than two. Specifically, in bin 3 (the bin with stronger biases), the number of biases against females is approximately ten times greater than the number of biases against males.

\subsubsection{Gender bias in Italian translation}
In this section, we will analyze gender bias in translations from English to Italian to assess its impact on the intensity and direction of bias.
In the next stage, we computed the \textit{GenderBiasIntensity} for Italian translations, as previously done with English. This numerical feature spans from $0.0$ to $0.26$. Notably, the maximum value within this range has decreased by $0.18$ compared to its English equivalent, indicating an overall reduction in gender bias intensity. Based on this range, we classified the words into three distinct bins, as illustrated in Table 5.

\begin{table}[ht]
\centering
\caption{Number of words per bin along with representative examples for comparing the intensity of bias in Italian translation.}
\label{tab:5}
\begin{tabular}{|c|c|c|c|}
\hline
\textbf{Bins} & \textbf{Range} & \textbf{Words Count} & \textbf{Examples}                                                                          \\ \hline
1             & 0.0 - 0.1      & 199                  & \begin{tabular}[c]{@{}c@{}}architetto, bellissimo, medico,\\ casalinga, rugby\end{tabular} \\ \hline
2 & 0.1 - 0.2 & 107 & \begin{tabular}[c]{@{}c@{}}ballerina, baseball, cosmetici, \\ dittatore, chitarrista\end{tabular} \\ \hline
3             & 0.2 - 0.3      & 27                   & \begin{tabular}[c]{@{}c@{}}zia, imprenditrice, calcio, \\ signora, infermiera\end{tabular} \\ \hline
\end{tabular}
\end{table}

In the Italian translation, similarly to English, the last bin (bin 3) comprises gender-specific words like \textit{zia (aunt), mamma (mom), lei (she), lui (he),} and \textit{donna (woman)}. Bins 1 and 2 are potential candidates for bias, where bin 2 exhibits the strongest bias intensity, ranging between $0.1$ and $0.2$.
Following the calculation of bias intensity, determining the bias direction is essential. 
The range of this feature for Italian translations spans from $-0.21$ to $0.26$. Subsequently, we classify it into 3 bins and 2 directions, as shown in Table 6.

\begin{table}[ht]
\centering
\caption{Number of words per bin for male and female direction for Italian translations.}
\label{tab:6}
\begin{tabular}{|cc|c|c|c|}
\hline
\multicolumn{1}{|c|}{\textbf{Bins}} & \textbf{Ranges} & \textbf{Female-directed} & \textbf{Male-directed} & \textbf{Total} \\ \hline
\multicolumn{1}{|c|}{1}  & 0.0 - 0.1 & 76           & 123          & \textbf{199} \\ \hline
\multicolumn{1}{|c|}{2}  & 0.1 - 0.2 & 52           & 55           & \textbf{107} \\ \hline
\multicolumn{1}{|c|}{3}  & 0.2 - 0.3 & 23           & 4            & \textbf{27}  \\ \hline
\multicolumn{2}{|c|}{\textbf{Total}} & \textbf{151} & \textbf{182} & \textbf{333} \\ \hline
\end{tabular}
\end{table}

Table 6 illustrates that although there is a reduction in the intensity of the bias after translation, there is a change in the direction of the bias. Specifically, while more than 69\% of the English words in Table 4 show a bias towards females, this ratio decreases to 45\% after translation. Another interesting observation is that after translation, the ratio of bias direction between females and males becomes more balanced, indicating an almost equal distribution.

\subsubsection{Post-translation Gender Shifts}
Although English is a notional gender language, we carried out a gender analysis of the words in the list before translation to determine the percentage of gender-specific and gender-neutral terms in the source data. Out of 333 words, 157 are gender neutral and 176 are gender specific (89 feminine and 87 masculine). An examination of the translations from Google Translate shows that of the 157 neutral English words, 91 are translated into masculine forms and only 32 into feminine forms. This ratio of approximately three to one provides a first insight into the gender bias of the model. This preliminary observation indicates a tendency for many English words to be translated into the masculine form in Italian, suggesting a potential gender bias in the translation model. Table 7 shows the details of this gender shift after translation, with some examples to illustrate this tendency. 
\begin{table}[ht]
\centering
\caption{Percentage of post-translation gender shifts with some examples.}
\label{tab:7}
\begin{tabular}{|ccc|ccc|c|}
\hline
\multicolumn{3}{|c|}{\textbf{\begin{tabular}[c]{@{}c@{}}Pre-translation \\ Genders (English)\end{tabular}}} &
  \multicolumn{3}{c|}{\textbf{\begin{tabular}[c]{@{}c@{}}Post-translation \\ Genders (Italian)\end{tabular}}} &
  \multirow{2}{*}{\textbf{Examples}} \\ \cline{1-6}
\multicolumn{1}{|c|}{\textbf{Gender}} &
  \multicolumn{1}{c|}{\textbf{Count}} &
  \textbf{\%} &
  \multicolumn{1}{c|}{\textbf{Gender}} &
  \multicolumn{1}{c|}{\textbf{Count}} &
  \textbf{\%} &
   \\ \hline
\multicolumn{1}{|c|}{\multirow{3}{*}{\textbf{Neutral}}} &
  \multicolumn{1}{c|}{\multirow{3}{*}{157}} &
  \multirow{3}{*}{47,2\%} &
  \multicolumn{1}{c|}{Neutral} &
  \multicolumn{1}{c|}{34} &
  10,20\% &
  adorable $\rightarrow$ adorabile \\ \cline{4-7} 
\multicolumn{1}{|c|}{} &
  \multicolumn{1}{c|}{} &
   &
  \multicolumn{1}{c|}{Masculine} &
  \multicolumn{1}{c|}{91} &
  27,32\% &
  architect $\rightarrow$ architetto \\ \cline{4-7} 
\multicolumn{1}{|c|}{} &
  \multicolumn{1}{c|}{} &
   &
  \multicolumn{1}{c|}{Feminine} &
  \multicolumn{1}{c|}{32} &
  9,60\% &
  homemaker $\rightarrow$ casalinga \\ \hline
\multicolumn{1}{|c|}{\multirow{3}{*}{\textbf{Masculine}}} &
  \multicolumn{1}{c|}{\multirow{3}{*}{87}} &
  \multirow{3}{*}{26,1\%} &
  \multicolumn{1}{c|}{Neutral} &
  \multicolumn{1}{c|}{6} &
  1,80\% &
  grandson $\rightarrow$ nipote \\ \cline{4-7} 
\multicolumn{1}{|c|}{} &
  \multicolumn{1}{c|}{} &
   &
  \multicolumn{1}{c|}{Masculine} &
  \multicolumn{1}{c|}{74} &
  22,22\% &
  actor $\rightarrow$ attore \\ \cline{4-7} 
\multicolumn{1}{|c|}{} &
  \multicolumn{1}{c|}{} &
   &
  \multicolumn{1}{c|}{Feminine} &
  \multicolumn{1}{c|}{7} &
  2,10\% &
  beard $\rightarrow$ barba \\ \hline
\multicolumn{1}{|c|}{\multirow{3}{*}{\textbf{Feminine}}} &
  \multicolumn{1}{c|}{\multirow{3}{*}{89}} &
  \multirow{3}{*}{26,7\%} &
  \multicolumn{1}{c|}{Neutral} &
  \multicolumn{1}{c|}{7} &
  2,10\% &
  spokeswoman $\rightarrow$ portavoce \\ \cline{4-7} 
\multicolumn{1}{|c|}{} &
  \multicolumn{1}{c|}{} &
   &
  \multicolumn{1}{c|}{Masculine} &
  \multicolumn{1}{c|}{13} &
  3,90\% &
  maternal $\rightarrow$ materno \\ \cline{4-7} 
\multicolumn{1}{|c|}{} &
  \multicolumn{1}{c|}{} &
   &
  \multicolumn{1}{c|}{Feminine} &
  \multicolumn{1}{c|}{69} &
  20.72\% &
  aunt $\rightarrow$ zia \\ \hline
\end{tabular}
\end{table}

To observe the effect of translation on similarity scores, we use the feature \textit{PostTranslationSimilarityChanges}. We computed this for pre- and post-translation words and the results are shown in Table 8. A positive value of this feature indicates an increase in the similarity score after translation. This increase is observed in 85.58\% of the words in the feminine version and in 96.39\% of the words in the masculine version. This indicates that translation generally leads to an increase in similarity scores. However, as this increase occurs in both directions (masculine and feminine), it does not increase the bias. From this analysis, it is clear that translation does not significantly affect bias intensity and that the bias is more related to the word embedding methods.

\begin{table}[ht]
\centering
\caption{The percentage and count of post-translation similarity changes.}
\label{tab:8}
\begin{tabular}{|cc|cc|cc|}
\hline
\multicolumn{1}{|c|}{\textbf{\begin{tabular}[c]{@{}c@{}}Pre-translation \\ Genders (English)\end{tabular}}} &
  \textbf{\begin{tabular}[c]{@{}c@{}}Post-translation \\ Genders (Italian)\end{tabular}} &
  \multicolumn{2}{c|}{\textbf{lei - she}} &
  \multicolumn{2}{c|}{\textbf{lui - he}} \\ \hline
\multicolumn{1}{|c|}{\textbf{Gender}} &
  \textbf{Gender} &
  \multicolumn{1}{c|}{\textbf{+}} &
  \textbf{-} &
  \multicolumn{1}{c|}{\textbf{+}} &
  \textbf{-} \\ \hline
\multicolumn{1}{|c|}{\multirow{3}{*}{\textbf{Neutral}}} &
  Neutral &
  \multicolumn{1}{c|}{30} &
  4 &
  \multicolumn{1}{c|}{34} &
  0 \\ \cline{2-6} 
\multicolumn{1}{|c|}{} &
  Masculine &
  \multicolumn{1}{c|}{78} &
  13 &
  \multicolumn{1}{c|}{90} &
  1 \\ \cline{2-6} 
\multicolumn{1}{|c|}{} &
  Feminine &
  \multicolumn{1}{c|}{28} &
  4 &
  \multicolumn{1}{c|}{32} &
  0 \\ \hline
\multicolumn{1}{|c|}{\multirow{3}{*}{\textbf{Masculine}}} &
  Neutral &
  \multicolumn{1}{c|}{5} &
  1 &
  \multicolumn{1}{c|}{5} &
  1 \\ \cline{2-6} 
\multicolumn{1}{|c|}{} &
  Masculine &
  \multicolumn{1}{c|}{66} &
  8 &
  \multicolumn{1}{c|}{68} &
  6 \\ \cline{2-6} 
\multicolumn{1}{|c|}{} &
  Feminine &
  \multicolumn{1}{c|}{6} &
  1 &
  \multicolumn{1}{c|}{6} &
  1 \\ \hline
\multicolumn{1}{|c|}{\multirow{3}{*}{\textbf{Feminine}}} &
  Neutral &
  \multicolumn{1}{c|}{4} &
  3 &
  \multicolumn{1}{c|}{7} &
  0 \\ \cline{2-6} 
\multicolumn{1}{|c|}{} &
  Masculine &
  \multicolumn{1}{c|}{11} &
  2 &
  \multicolumn{1}{c|}{12} &
  1 \\ \cline{2-6} 
\multicolumn{1}{|c|}{} &
  Feminine &
  \multicolumn{1}{c|}{57} &
  12 &
  \multicolumn{1}{c|}{67} &
  2 \\ \hline
\multicolumn{2}{|c|}{\textbf{Total}} &
  \multicolumn{1}{c|}{\textbf{285}} &
  \textbf{48} &
  \multicolumn{1}{c|}{\textbf{321}} &
  \textbf{12} \\ \hline
\multicolumn{2}{|c|}{\textbf{Percentage}} &
  \multicolumn{1}{c|}{\textbf{85,58\%}} &
  \textbf{14,41\%} &
  \multicolumn{1}{c|}{\textbf{96,39\%}} &
  \textbf{3,60\%} \\ \hline
\end{tabular}
\end{table}

Based on the above observation, we can conclude that to achieve unbiased translation, we must first reduce bias in the word embedding methods used to train the MT models. Unbiased translation, particularly in the context of gender, requires minimizing the introduction or reinforcement of stereotypes. As discussed in the previous section, one challenge in dealing with grammatically gendered languages such as Italian is the default use of masculine forms. Word embeddings in Italian, as in many languages, show significant gender bias. One study found that although Italian word embeddings have less potential to reinforce certain stereotypes than English, the presence of grammatical gender introduces different forms of bias. For example, in job search contexts, masculine terms may be the default, potentially disadvantaging women by making male candidates more likely to be found~\cite{Biasion2020}.

\section{Conclusion}

Our work has provided us with several findings that help to answer the research questions we have identified.

In response to RQ1, the overall conclusion is that MT still affects the use of gender in texts today. However, the translation step per se does not have such a negative impact on bias. Most of the bias comes from the word embedding methods used in the training of MT algorithms. These methods are heavily influenced by the distribution of words in a language. This leads us to RQ2. If the bias is mainly related to the distribution of words in the source and target languages, then the problems of stereotypes in MT translation are less related to the algorithm itself and more related to stereotypical language spillovers in the corpora used for training~\cite{savoldi2021gender}. This conclusion invites further studies on how to efficiently structure and control both training and control datasets. 

To address RQ3, a notable aspect of our research is the investigation of the impact of translation on bias, using the widely used Google Translate. Our study integrates embedding and translation, which is a unique approach. Compared to previous literature, which has mainly focused on \textit{Word2Vec}~\cite{bolukbasi2016man}, we pay particular attention to embedding methods such as \textit{FastText}, which are better suited for multilingual analysis. Furthermore, while much research in this area has focused only on lists of occupational terms~\cite{Biasion2020-yb, stanovsky2019evaluating, prates2020assessing}, our analysis uses a comprehensive list of words that includes different occupations, adjectives, sports, and more. This approach allows us to identify bias in all aspects of language, not just job titles.

Nevertheless, regarding RQ3, our study advances the state of the art in quantifying gender bias in MT by using numerical features to effectively assess the intensity and direction of bias. Our approach not only identifies the presence of bias but also provides a normalization measure to identify its magnitude and directional tendencies, allowing for a more precise assessment compared to previous studies. This paves the way for the application of our methodology to different languages and MT models, thereby broadening its applicability and impact in the field.
Furthermore, by highlighting specific areas where bias is most pronounced, we provide actionable guidance for researchers and practitioners seeking to improve the fairness of their MT systems. In our observations, we find a bias in the \textit{FastText} embedding method towards both males and females, as evidenced by the significant deviation of many words in the scatter plot from the diagonal line. The range of related numerical features confirms this bias even if it is not particularly severe.
While most words have a greater similarity to \textit{she} than to \textit{he}, indicating a bias toward females, the overall direction of the bias shifts after translation. After translation, there is a reduction in the intensity of the bias, with \textit{GenderBiasIntensity} ranging from $0.0$ to $0.26$. However, there is a significant change in the direction of the bias, with many words becoming more male-biased. This suggests that while \textit{FastText} has a female bias, \textit{Google Translate} often translates words into the masculine form, significantly changing the direction of the bias.

The limitations of our work are the use of only one MT and the embedding model for analysis. Thus, future research could explore other word embedding methods, such as \textit{GloVe}, \textit{Bag-of-Words}, or pre-trained models, in conjunction with different translation models to determine which combinations produce less gender-biased results. Another limitation of our work is that the detection of stereotypes and biases is challenging from a linguistic perspective. According to the definitions of 'bias' that we have adopted, it is difficult to detect them automatically, as it requires a lot of cultural context to understand and properly detect them without a full-fledged ontology. Therefore, further studies could focus on fine-tuning LLMs to better detect stereotypes and biases by exploring the possibilities of studying the taxonomy and ontology of stereotypes and biases in a target language, as some have already started to do ~\cite{doughman-etal-2021-gender}. Presenting some methods to mitigate the bias in languages such as Italian is also an interesting topic for future directions. Finally, in this study, we have only focused on gender bias. The study of other forms of bias, such as age, nationality, race, and religion, therefore provides a broad direction for further research. Comparative analyses of these models would provide valuable insights into their relative biases and pave the way for future research and refinement.

\appendix 
\section*{Acknowledgments}
 The work reported in this paper has been partly funded by the European Union - NextGenerationEU, under the National Recovery and Resilience Plan (NRRP) Mission 4 Component 2 Investment Line 1.5: Strengthening of research structures and creation of R\&D “innovation ecosystems”, set up of “territorial leaders in R\&D”, within the project “MUSA - Multilayered Urban Sustainability Action” (contract n. ECS 00000037).

\end{document}